\documentclass{article}
\usepackage{graphicx} 
\usepackage{url} 

\title{AI Runtime Infrastructure.}
\author{Christopher Cruz}
\date{February 2026}

\begin{document}

\maketitle

\begin{abstract}
Agentic AI systems increasingly operate over long horizons, interact with external tools, and must adapt to dynamic environments during execution. While significant progress has been made in model serving infrastructure, orchestration frameworks, and post-hoc observability, these approaches do not address failures, inefficiencies, and safety risks that emerge \emph{during} agent execution. In practice, many of the most costly agent failures occur at runtime, after planning has begun and outside the scope of static orchestration or offline analysis.

We introduce \textbf{AI Runtime Infrastructure}, a distinct execution-time layer that operates above the model and below the application, actively observing, reasoning over, and intervening in agent behavior to optimize task success, latency, token efficiency, reliability, and safety while the agent is running. Unlike model-level optimizations or passive logging systems, runtime infrastructure treats execution itself as an optimization surface, enabling adaptive memory management, failure detection, recovery, and policy enforcement over long-horizon agent workflows.

We formalize the scope, responsibilities, and boundaries of AI runtime infrastructure, distinguishing it from related areas such as inference optimization, agent orchestration, and observability tooling. We outline core design principles for runtime systems, including execution-time intervention, long-horizon state awareness, and integrated recovery mechanisms. Finally, we describe Adaptive Focus Memory (AFM) and VIGIL as early instantiations of this layer, demonstrating how runtime infrastructure can materially improve agent robustness and efficiency in real-world settings.

We argue that AI runtime infrastructure represents a foundational component of scalable and reliable agentic systems, and that formalizing this layer is necessary for the next generation of production-grade AI agents.

\end{abstract}
\section{Introduction}

Agentic AI systems are increasingly deployed to perform complex tasks over long horizons, interacting with external tools, APIs, and environments while operating under latency, cost, and safety constraints. Unlike single-turn model inference, these systems execute multi-step workflows in which decisions made early in execution can have cascading effects on downstream behavior, resource consumption, and failure modes. As a result, many of the most significant challenges in production agentic systems arise not at model invocation time, but during execution itself.

Existing infrastructure has largely addressed adjacent concerns. Model serving and inference infrastructure focuses on optimizing the execution of individual model calls through techniques such as batching, caching, and hardware-aware scheduling. Agent orchestration frameworks provide abstractions for composing tools, prompts, and control flow, enabling developers to specify how agents should act. Observability and AgentOps tooling captures logs, traces, and metrics to support debugging and offline analysis. Safety mechanisms are often applied post-hoc, filtering or moderating outputs after generation. While each of these layers is essential, none is designed to actively intervene in agent behavior \emph{during} execution.

In practice, agent failures frequently emerge after execution has begun: context windows overflow, intermediate reasoning drifts off task, tool interactions compound errors, or latent safety risks surface mid-workflow. Because these failures occur at runtime, static orchestration logic and post-hoc analysis are insufficient to prevent or mitigate them. Once an agent has entered an unrecoverable execution state, logging the failure provides insight but does not restore correctness, efficiency, or safety.

This gap suggests the need for a distinct execution-time layer that treats agent runtime behavior itself as a first-class optimization surface. Such a layer must be capable of observing execution state over long horizons, reasoning about emerging failure modes, and intervening dynamically to adjust memory, control flow, resource usage, or policy enforcement while the agent is running. Importantly, this functionality is orthogonal to model-level optimization and application-specific logic, and cannot be reduced to either.

In this work, we introduce \textbf{AI Runtime Infrastructure}, a systems layer that operates above the model and below the application, providing active execution-time oversight and intervention for agentic systems. We argue that formalizing this layer is necessary for building scalable, reliable, and safe AI agents, and that its absence represents a structural limitation in current agent deployments. The remainder of this paper defines the scope and boundaries of AI runtime infrastructure, situates it relative to prior work, and describes early instantiations that demonstrate its practical value.

\section{Defining AI Runtime Infrastructure}

We define \textbf{AI Runtime Infrastructure} as an execution-time systems layer that actively observes, reasons over, and intervenes in the behavior of agentic AI systems while they are running. This layer operates above the model and below the application, and is responsible for optimizing agent execution with respect to task success, latency, token efficiency, reliability, and safety over long horizons.

Unlike model serving infrastructure, which focuses on optimizing the performance of individual inference calls, AI runtime infrastructure treats agent execution itself as a first-class object. Its scope includes monitoring evolving execution state, detecting emerging failure modes, and applying corrective actions during runtime rather than after execution has completed. This distinction is critical in agentic systems, where errors often compound across steps and cannot be addressed through static orchestration or post-hoc analysis alone.

Formally, an AI runtime infrastructure system satisfies three necessary properties. First, it operates \emph{during execution}, maintaining continuous visibility into agent state, intermediate outputs, and environmental interactions across multiple steps. Second, it performs \emph{active intervention}, modifying execution behavior through actions such as adaptive memory management, control-flow adjustment, recovery triggering, or policy enforcement. Third, it reasons over \emph{long-horizon context}, incorporating execution history rather than relying solely on the current prompt or model invocation.

AI runtime infrastructure is distinct from several adjacent categories of systems. It does not encompass inference optimization techniques such as caching, batching, or hardware-aware scheduling, which improve model execution but do not reason about agent behavior. It is not equivalent to agent orchestration frameworks, which define control flow and tool composition but lack execution-time introspection and adaptive intervention. It is also separate from observability and AgentOps tooling, which capture execution traces for offline analysis but do not influence outcomes while an agent is running. Finally, runtime infrastructure differs from post-hoc safety layers, as it addresses safety risks as they emerge during execution rather than filtering outputs after generation.

The responsibilities of AI runtime infrastructure include, but are not limited to: maintaining execution-time state representations; identifying deviations from task objectives or safety constraints; allocating and compressing contextual information; triggering recovery or rollback mechanisms; and enforcing runtime policies that balance efficiency, robustness, and risk. Importantly, this layer is designed to be application-agnostic, providing general-purpose execution control rather than encoding domain-specific logic.

By formalizing AI runtime infrastructure as a distinct systems layer, we aim to clarify the architectural requirements of reliable agentic AI and to provide a foundation for principled design and evaluation. The following sections situate this layer within the broader AI systems stack and outline core design principles for effective runtime infrastructure.

\section{Architectural Positioning}

\begin{figure}[t]
\centering
\includegraphics[width=\linewidth]{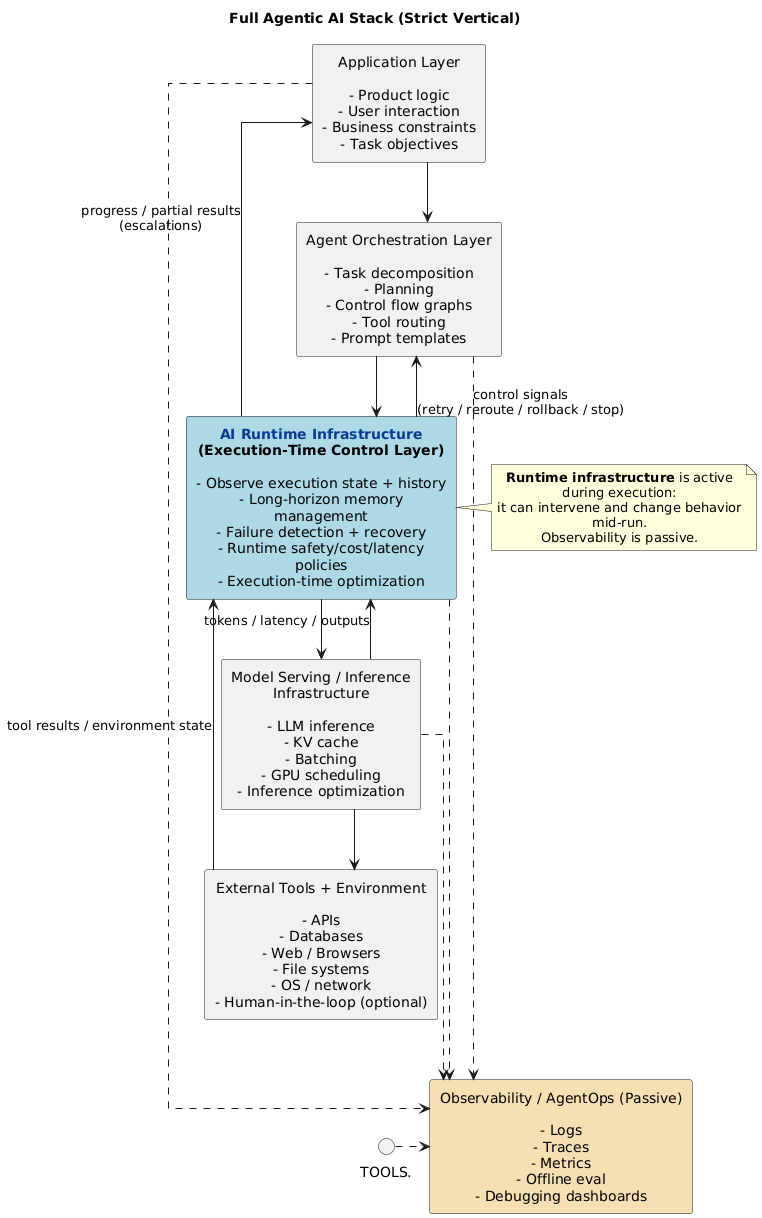}
\caption{The full agentic AI systems stack. AI runtime infrastructure operates as an execution-time control layer between agent orchestration and model serving, observing execution state and intervening during runtime to optimize task success, efficiency, reliability, and safety. Observability and evaluation systems span the stack but do not influence execution-time behavior.}
\label{fig:airinfra-stack}
\end{figure}

AI runtime infrastructure occupies a distinct position within the \emph{agentic systems stack}, operating between model execution and application-level logic. This placement is intentional: runtime infrastructure must have sufficient proximity to the model to observe intermediate outputs and resource utilization, while remaining abstracted from application-specific objectives and domain logic. Figure~\ref{fig:airinfra-stack} illustrates the full agentic systems stack and the architectural role of AI runtime infrastructure within it.

At the lowest level of the stack, model serving and inference infrastructure is responsible for executing individual model calls efficiently. This includes concerns such as batching, caching, hardware scheduling, and latency optimization. These components expose inference capabilities but do not reason about multi-step agent behavior or execution history.

Above this layer, AI runtime infrastructure maintains continuous visibility into the agent's execution state across steps. It consumes signals such as intermediate model outputs, tool responses, memory utilization, and policy constraints, and uses these signals to make execution-time decisions. Crucially, this layer is empowered to intervene during execution, for example by modifying contextual inputs, adjusting control flow, triggering recovery mechanisms, or enforcing runtime policies. These interventions occur without requiring changes to the underlying model or to the application logic that invokes the agent.

At the top of the stack, application logic specifies task objectives, user interaction patterns, and domain-specific behavior. Applications define what an agent should accomplish, but they typically lack mechanisms to monitor or correct execution failures as they unfold. By decoupling execution oversight from application logic, AI runtime infrastructure enables reusable, application-agnostic control over agent behavior.

Architecturally, AI runtime infrastructure can be implemented as an execution-time control plane that interfaces with both the agent execution loop and external system resources. The agent produces execution artifacts—such as intermediate reasoning steps, tool invocations, and partial outputs—which are observed by the runtime layer. In response, the runtime layer may emit control signals that alter subsequent execution, forming a closed feedback loop that persists for the duration of the agent's operation. This structure distinguishes runtime infrastructure from static orchestration pipelines, which define execution paths but do not adapt based on observed outcomes.

Importantly, AI runtime infrastructure does not replace existing layers but composes with them. Model serving infrastructure remains responsible for efficient inference, orchestration frameworks continue to manage high-level task decomposition, and observability systems provide retrospective analysis without influencing execution-time behavior. Runtime infrastructure complements these components by providing execution-time intelligence that bridges the gap between planning and outcome. By explicitly formalizing this architectural role, we clarify how adaptive, reliable, and safe agentic systems can be constructed without entangling concerns across layers.

\section{Design Principles for AI Runtime Infrastructure}

AI runtime infrastructure introduces a distinct set of design requirements that differ from those of model serving systems, orchestration frameworks, and observability tooling. To clarify what constitutes effective runtime infrastructure for agentic systems, we outline a set of core design principles. These principles are not tied to specific implementations, but instead characterize the essential properties required for execution-time oversight and control.

\subsection{Execution-Time Intervention}

AI runtime infrastructure must be capable of intervening \emph{during} agent execution rather than operating solely before or after a run. Many agent failures emerge only after execution has begun, when intermediate reasoning, tool interactions, or accumulated context diverge from intended objectives. Systems that observe failures but cannot alter execution behavior in response do not satisfy this requirement. Runtime infrastructure must therefore support mechanisms that can modify agent inputs, control flow, or execution state while the agent is actively running.

\subsection{Long-Horizon State Awareness}

Agentic systems frequently operate over extended horizons involving dozens or hundreds of steps. AI runtime infrastructure must maintain visibility into execution history across these horizons, rather than relying exclusively on the current prompt or most recent model output. This includes tracking intermediate decisions, memory utilization, tool outcomes, and prior interventions. Without long-horizon state awareness, runtime systems are unable to reason about cumulative failure modes or compounding inefficiencies.

\subsection{Closed-Loop Control}

Effective runtime infrastructure forms a closed feedback loop between observation and action. Execution signals produced by the agent—such as intermediate outputs, latency measurements, or tool responses—are continuously evaluated and used to inform subsequent interventions. This closed-loop structure distinguishes runtime infrastructure from static orchestration pipelines, which define execution paths in advance but do not adapt based on observed outcomes during execution.

\subsection{Model-Agnostic Operation}

AI runtime infrastructure should operate independently of specific model architectures or providers. While it must interface closely with model execution to observe outputs and resource usage, it should not require modification of the underlying model or rely on model-specific internals. This separation enables runtime infrastructure to generalize across different models and to evolve independently as model capabilities change.

\subsection{Application-Agnostic Control}

Runtime infrastructure is designed to provide execution-time control that is reusable across applications. It should not encode domain-specific task logic or application-level objectives, which remain the responsibility of the application layer. By maintaining this separation, runtime infrastructure can serve as a general-purpose control plane that supports diverse agentic workloads without entangling execution oversight with business logic.

\subsection{Safety, Cost, and Reliability as Runtime Concerns}

Safety, efficiency, and reliability constraints must be enforced as part of execution-time decision making rather than solely through static policies or post-hoc filtering. AI runtime infrastructure enables these concerns to be evaluated dynamically as execution unfolds, allowing systems to respond to emerging risks, escalating costs, or degraded performance before failures become irreversible. Treating these dimensions as runtime concerns is essential for deploying agentic systems in production environments.

Together, these principles define AI runtime infrastructure as an execution-time control layer that complements existing components of the agentic systems stack. Systems that satisfy these criteria can actively shape agent behavior as it unfolds, enabling adaptive, robust, and scalable agentic AI beyond what static orchestration or offline analysis alone can provide.

\section{Early Systems and Precursors}

The formalization of AI runtime infrastructure is motivated by practical challenges encountered in long-horizon agentic systems, where failures, inefficiencies, and safety risks emerge during execution rather than at planning time. Prior to the explicit definition of runtime infrastructure as an execution-time control layer, several systems addressed aspects of runtime behavior without fully satisfying the criteria outlined in Section 4. In this section, we describe two such systems—VIGIL and Adaptive Focus Memory (AFM)—to illustrate the progression from runtime-aware precursors to a fully realized instantiation of AI runtime infrastructure.

\subsection{VIGIL: A Runtime-Aware Precursor}

VIGIL~\cite{vigil} is a reflective runtime system designed to diagnose and respond to failures in long-running agent workflows. It analyzes structured execution logs and traces to detect anomalous behavior, degraded performance, or violations of expected execution patterns, and can trigger remediation actions or human-in-the-loop escalation. By reasoning over execution histories that span many agent steps, VIGIL demonstrates the limitations of purely post-hoc observability in agentic systems.

While VIGIL is explicitly runtime-aware, it operates primarily outside the agent execution loop. Its diagnostic and recovery mechanisms are invoked after failures have been detected, and its influence on agent behavior occurs through external remediation rather than continuous, in-loop control. As a result, VIGIL does not perform execution-time intervention in the sense required for AI runtime infrastructure. Instead, it serves as a precursor system that exposes the need for tighter integration between execution monitoring and control, and motivates the development of runtime infrastructure capable of intervening directly during agent execution.

\subsection{Adaptive Focus Memory: AI Runtime Infrastructure}

Adaptive Focus Memory (AFM)~\cite{afm} represents an early instantiation of AI runtime infrastructure as defined in this work. AFM operates directly within the agent execution loop, continuously observing execution state and intervening in real time to manage contextual information over long horizons. By dynamically allocating, compressing, and reweighting memory during execution, AFM actively shapes agent behavior while tasks are in progress.

AFM satisfies the core properties of AI runtime infrastructure. It performs execution-time intervention by modifying the contextual inputs provided to the model as execution unfolds. It reasons over long-horizon state by maintaining and adapting memory representations across many agent steps. Finally, it participates in a closed-loop control process, where execution signals inform subsequent interventions that directly influence agent behavior. These operations occur without requiring changes to the underlying model or application logic, positioning AFM as an execution-time control layer rather than an orchestration or observability component.

\subsection{From Precursors to Runtime Infrastructure}

Together, VIGIL and AFM illustrate the evolution from runtime-aware monitoring toward fully integrated execution-time control. VIGIL demonstrates that post-hoc diagnostics and recovery are insufficient for managing long-horizon agent failures, while AFM operationalizes the principles of AI runtime infrastructure by embedding adaptive control directly into agent execution. This progression underscores the necessity of formalizing runtime infrastructure as a distinct systems layer and clarifies the architectural and functional boundary between precursors and true execution-time control systems.

\section{Related Work}

AI runtime infrastructure intersects with several established areas of research and engineering, including model serving infrastructure, agent orchestration frameworks, observability and AgentOps tooling, and AI safety systems. While these domains address important aspects of agentic system deployment, they do not provide execution-time control over agent behavior as defined in this work.

\subsection{Model Serving and Inference Infrastructure}

A large body of work focuses on optimizing the execution of individual model invocations through techniques such as batching, caching, quantization, and hardware-aware scheduling. These systems aim to improve throughput, latency, and cost efficiency at inference time, and are critical for deploying large-scale language models in production environments. However, model serving infrastructure treats each inference call largely in isolation and does not reason about multi-step agent execution, long-horizon state, or task-level outcomes. As a result, inference optimization alone is insufficient for managing failures or inefficiencies that emerge during extended agent workflows.

\subsection{Agent Orchestration Frameworks}

Agent orchestration frameworks provide abstractions for composing prompts, tools, and control flow into structured agent behaviors. These frameworks enable developers to specify execution graphs, routing logic, and tool usage patterns, and have been instrumental in accelerating the development of agentic systems. However, orchestration frameworks primarily define execution \emph{plans} rather than execution-time control. Once an agent is running, orchestration logic typically executes as specified, with limited ability to adapt based on observed runtime behavior. In contrast, AI runtime infrastructure reasons over execution state as it unfolds and intervenes dynamically to influence agent behavior during operation.

\subsection{Observability and AgentOps Tooling}

Observability and AgentOps systems capture logs, traces, metrics, and evaluation artifacts from agent executions to support debugging, monitoring, and offline analysis. These tools provide valuable insight into agent performance and failure modes, particularly in production settings. However, they are inherently retrospective: execution data is collected for inspection after the fact, and does not directly influence agent behavior during execution. While runtime-aware precursors such as VIGIL~\cite{vigil} demonstrate the limitations of purely post-hoc analysis, observability tooling alone does not satisfy the requirements of execution-time intervention and closed-loop control.

\subsection{AI Safety and Policy Enforcement Systems}

AI safety mechanisms are often implemented as static policies or post-hoc filters that constrain or moderate model outputs. These approaches play an important role in mitigating harmful behavior, but typically operate outside the agent execution loop and lack visibility into long-horizon execution state. More recent work explores adaptive safety mechanisms that respond to contextual signals, but these are rarely integrated as general-purpose execution-time control layers. AI runtime infrastructure treats safety, reliability, and efficiency as runtime concerns, enabling dynamic intervention as risks emerge during execution rather than solely at output time.

\subsection{Positioning AI Runtime Infrastructure}

AI runtime infrastructure complements, rather than replaces, these existing systems. Model serving infrastructure remains responsible for efficient inference, orchestration frameworks continue to define high-level behavior, observability tooling supports retrospective analysis, and safety systems enforce constraints. Runtime infrastructure addresses a distinct gap by providing execution-time oversight and control across long-horizon agent workflows. By formalizing this layer, we clarify architectural boundaries and enable principled design of systems that actively shape agent behavior as it unfolds.

\section{Implications and Future Directions}

Formalizing AI runtime infrastructure as a distinct execution-time layer has several implications for the design, evaluation, and deployment of agentic systems. By treating agent execution itself as an optimization surface, runtime infrastructure enables new classes of adaptive behavior that are difficult or impossible to achieve through static orchestration, model-level optimization, or post-hoc analysis alone.

\subsection{Scalable Reliability for Long-Horizon Agents}

As agentic systems are deployed to perform increasingly long-horizon tasks, failure modes that compound over time become a dominant source of cost and unreliability. AI runtime infrastructure provides a mechanism for addressing these failures during execution, before they propagate into irrecoverable states. This suggests a shift from reactive debugging toward proactive execution-time control as a foundation for scalable agent reliability.

\subsection{Runtime-Aware Safety and Governance}

Treating safety and policy enforcement as runtime concerns enables more nuanced and adaptive governance of agent behavior. Rather than relying solely on static constraints or output filtering, runtime infrastructure can respond dynamically to evolving execution context, emerging risks, or changes in environmental conditions. This opens opportunities for safety mechanisms that are sensitive to long-horizon behavior and cumulative risk, rather than isolated model outputs.

\subsection{Evaluation Beyond Post-Hoc Metrics}

The presence of an execution-time control layer also motivates new approaches to evaluating agentic systems. Traditional metrics that summarize outcomes after execution may fail to capture the benefits of runtime intervention, such as avoided failures or reduced recovery costs. Future evaluation frameworks may need to account for execution trajectories, intervention timing, and counterfactual outcomes enabled by runtime infrastructure.

\subsection{Open Research Directions}

AI runtime infrastructure introduces several open research challenges. These include designing principled policies for intervention under uncertainty, balancing competing objectives such as efficiency and safety at runtime, and developing abstractions that generalize across diverse agent architectures and environments. Additionally, understanding how runtime infrastructure interacts with learning-based adaptation remains an open question, particularly in systems that combine execution-time control with online or continual learning.

More broadly, formalizing runtime infrastructure highlights the need for clearer architectural boundaries in agentic AI systems. As agents become more autonomous and are entrusted with higher-impact tasks, execution-time control is likely to become a foundational requirement rather than an optional enhancement.

\section{Conclusion}

Agentic AI systems increasingly operate over long horizons, interact with external environments, and must satisfy constraints on reliability, efficiency, and safety during execution. While existing infrastructure addresses model execution, orchestration, observability, and post-hoc evaluation, these components do not provide execution-time control over agent behavior. As a result, many critical failure modes remain unaddressed until after execution has already degraded or failed.

In this work, we formalized \emph{AI runtime infrastructure} as a distinct execution-time systems layer that operates above the model and below the application. We defined its scope, responsibilities, and architectural boundaries, and identified core design principles that distinguish runtime infrastructure from adjacent systems. Through the examination of runtime-aware precursors and early instantiations, we illustrated how execution-time intervention enables adaptive control that cannot be achieved through static orchestration or retrospective analysis alone.

By explicitly naming and formalizing this layer, we aim to clarify the architectural requirements of scalable, reliable agentic systems and to provide a foundation for principled system design and evaluation. As agentic AI continues to move toward more autonomous and high-impact deployments, execution-time control is likely to become a foundational requirement rather than an optional enhancement. AI runtime infrastructure provides a framework for meeting this requirement and for advancing the next generation of production-grade agentic systems.

\end{document}